%% file: pai.tex
\documentclass[letterpaper,11pt,reqno]{amsart}

\input{MathMacros}
\usepackage{hyperref}
\usepackage{dirtytalk}
\usepackage{xpatch}

\begin{document}

\author[Reuben Brasher,
Nat Roth,
Justin Wagle
]{%
Reuben Brasher,
Nat Roth,
Justin Wagle\\
\MakeLowercase{\{rebrashe, narot, justiwag\}@microsoft.com
}}

\title[sometimes you want to go where everybody knows your name]{sometimes you want to go where everybody knows your name\\
\tiny{a metric for personalization}}
\date{\today}

\maketitle

\newcommand{\Loss}{\mathcal{L}}

\begin{abstract}
We introduce a new metric for measuring how well a model personalizes to a user's specific preferences. We define personalization as a weighting between performance on user specific data and performance on a more general global dataset that represents many different users. This global term serves as a form of regularization that forces us to not overfit to individual users who have small amounts of data. In order to protect user privacy, we add the constraint that we may not centralize or share user data. We also contribute a simple experiment in which we simulate classifying sentiment for users with very distinct vocabularies. This experiment functions as an example of the tension between doing well globally on all users, and doing well on any specific individual user. It also provides a concrete example of how to employ our new metric to help reason about and resolve this tension. We hope this work can help frame and ground future work into personalization.

\end{abstract}

\section{Introduction}

In a wide range of fields, from music and advertising recommendations to healthcare and a wide range of other consumer applications, learning users' personal tendencies and judgements is essential. Many current approaches demand centralized data storage and computation to aggregate and learn globally. Such central models, along with features known about a given user, make predictions appear personal to that user. While such global models have proven to be widely effective, they bring with them inherent conflicts with privacy. User data must leave the device, and training a central model requires regular communication between a given user and the remote model. Further, if users are in some way truly unique, and exhibit difference preferences than seemingly similar users, large centralized models may have trouble quickly adapting to this behavior.

With these disadvantages in mind, we present a definition of personalization that allows for no direct sharing or centralization of user data. We see personalization as the balance between generalization to global information and  specialization to a given user's quirks and biases.

To make this definition concrete, we show how a simple baseline model's performance changes on a sentiment analysis task as a function of user bias, and the way information is shared across models. We hope this work can contribute to framing the discussion around personalization and provide a metric for evaluating in what ways a model is truly providing a user personal recommendations.  

We also discuss related areas such as differential privacy, and federated learning, which have been motivated by similar considerations. Our work could easily fit into the frameworks of federated learning or differential privacy.

\section{Related Work}

\subsection{Personalized Models}
There has been a long history of research into personalization within machine learning. There is a wealth of work on using Bayesian hierarchical models to learn mixes of user and global parameters from data. These works have achieved success in areas from health care \cite{fan2015bayesian}, to recommendation systems \cite{zhang2007efficient}, to generally dealing with a mix of implicit and explicit feedback \cite{zigoris2006bayesian}. There has also been increasing work on helping practitioners to integrate these Bayesian techniques with deep learning models \cite{shi2017zhusuan}.

Many approaches to personalization within deep learning have relied on combining personal features, hand written or learned, with some more global features to make predictions. For example, in deep recommender systems, a feature might whether a user is a certain gender, or has seen a certain movie. A deep model may learn to embed these features, and combine them with some linear model as in \cite{cheng2016wide} in order to make recommendations for a specific user. It is also common to learn some vector describing the user end to end for a task, rather than doing this featurization by hand. In such scenarios your input might be a sentence and a user id and the prediction would be the next sentence as in \cite{al2016conversational}, in which the user is featurized via some learned vector. Similarly, Park et al \cite{park2017attend}, learn a vector representation of a user's context to generate image captions that are personal to the user, and \cite{chen2017user} learn a user vector alongside a language model to determine if a set of answers to a question will satisfy a user. These approaches have the benefit of not requiring any manual description of the important traits of a user. 

Here, when we discuss personalization, we focus more on personalization work within deep learning. In general, deep learning models are large, complicated, and very non-linear. This makes it hard to reason about how incorporating a new user, or set of training examples will affect the state of the model at large, a phenomenon known as \say{catastrophic forgetting} \cite{french1999catastrophic}, a topic which itself has seen a large amount of research \cite{kirkpatrick2017overcoming}, \cite{kemker2017measuring}, \cite{aljundi2017memory}. In general, this means that if we add a new user whose behavior is very different from our previous training examples, we need to take extra steps to preserve our performance on previous users. This makes online personalization of models to outlier users an open problem within deep learning.

\subsection{Federated Learning}
Our other key personalization constraint is privacy related; to get users to trust a model with extremely personal data, it is our believe that it is becoming increasingly necessary, and even legally mandated, to guarantee them a degree of privacy \cite{GGDPR}, \cite{CaliforniaPrivacy}. Research on federated learning has demonstrated that intelligence from users can be aggregated and centralized models trained without ever directly storing user data in a central location, alleviating part of these privacy concerns. This research focuses on training models when data is distributed on a very large number of devices, and further assumes each device does not have access to a representative sample of the global data \cite{mcmahan2016federated}, \cite{konevcny2016federated}. We therefore believe federating learning is a key part of any personalization strategy. 

Federated learning is concerned with training a central model that does well on users globally. However, the contribution from an individual user tends to be washed out after each update to the global model. Konecny et al., \cite{konevcny2016federated} admit as much, explicitly saying that the issues of personalization are separate from federated learning. Instead, much of the current research focus on improving communication speed \cite{konevcny2016federatedcommunication} and how to maintain stability between models when communication completely drops or lags greatly \cite{smith2017federated}. \cite{malle2017more} comes closest to our concerns, as it hypothesizes a system in which each user has a personal set of knowledge and some more global mechanism aggregating knowledge from similar users. They do not propose an exact mechanism for how to do this aggregating and how to determine which users are similar. We hope to contribute to the conversation on how to best minimally compromise the privacy and decentralization of learning, while not enforcing all models to globally cohere and synchronize. 

Finally, it is important to note that federation itself does not guarantee privacy. While in practice this aggregation of gradients, in the place of storing of raw data, will often obscure some user behavior, it may still leak information about users. For example, if an attacker observes a non-zero gradient for a feature representing a location, it may be trivial to infer that some of the users in the group live in that location. Making strong guarantees about the extent to which data gives us information about individual users is the domain of differential privacy \cite{dwork2008differential}, \cite{abadi2016deep}. In future work, we hope to incorporate these stronger notions of privacy into our discussion as well, but believe that federated learning is a good first step towards greater user privacy.

\section{Personalization Definition}

With these problems in mind, we define personalization as the relative weighting between performance of a model on a large, multi-user, global dataset, and the performance of that model on data from a single user. This definition implies several things. In particular, the extent to which a model can be personalized depends both on the model itself, and the spread of user behavior. On a task in which users always behave the same, there is little room for personalization, as a global model trained on all user data will likely be optimal globally and locally. However, on any task where user behavior varies significantly between individuals, it is possible a model trained on all users may perform poorly on any specific user. Nonetheless, a specific user may benefit from some global data; for example, a user with less training data may see better performance if they use a model trained with some global data. Therefore, the best personalization strategy will have some ability to incorporate global knowledge, will minimally distorting the predictions for a given user.

In addition, we add the constraint that user specific data be private to a user, and cannot be explicitly shared between models. In particular, this means that even if all user data is drawn from the same distribution, we cannot simply train on all the data. Instead we must determine other ways to share this knowledge, such as federating or ensembling user models.

In this paper, we establish some simple benchmarks for evaluating how well a model respects this definition of personalization.

Formally, suppose we have number of users, $N$, and for each user we have some user specific data, $\set{X_i}{i=1 \dots N}$, and user specific models, $\set{M_i}{i=1\dots N}$. Let the global data be $D = \bigcup X_i$, and we suppose we have a loss function, $\Loss$ , which is a function of both $X_i$ and $M_i$, $\Loss_i = \Loss \prn{X_i,\, M_i}$. We define our success at personalization as: 
\begin{equation}\label{eq:loss}
\alpha \Loss \prn{X_i,\, M_i} + \prn{1-\alpha} \Loss \prn{D,\, M_i}, 
\end{equation}
where $\alpha$ is between 0 and 1, and determines how much we weight local user and global data performance. In the case where $X_i$ follows the same distribution as all $D$, this definition trivially collapses to approximately optimizing $\Loss \prn{D,\, M_i}$, the familiar, non personal objective function on a dataset. However, as $\alpha$ increases and $X_i$ diverges from $D$, we introduce a tension between optimizing for the specific user, while still not ignoring the whole dataset. Finally, to enforce our definition of privacy, each model, $M_i$, has access only to $X_i$, and the weights of all the other models, $M_j$ for $j=1\dots N$, but does not have access to the other datasets, $M_j$, $j=1\dots N$, $j \ne i$.

\section{Personalization Motivation And Implications}

One question might be why we bother at all with adding global data to the equation, since it is more intuitive to think about personalization as just using the model that does best on a single user's data, and that data alone. However, that intuition ignores the fact that we may have only observed a small amount of behavior from any given user. If we only fit optimally to a specific user's data, we risk overfitting and performing poorly on new data even from that same user.

A Bayesian interpretation of our definition is to view the global data term as representing our prior belief about user behavior. Another interpretation is to view $\alpha$ as how much \say{catastrophic forgetting} we will allow our model to do in order to personalize to a user.

From the Bayesian perspective, the global data serves as a type of regularization that penalizes the local model from moving too far away from prior user data in order to fit to a new user. We can think about $\alpha$ as a hyperparameter representing the strength of our prior belief. The smaller $\alpha$ is, the less we allow the model to deviate from the global state. There may be no perfect rule for choosing $\alpha$, as it may depend on task, and rate at which we want to adapt to the user. 

One strategy could be to slowly increase $\alpha$ for a given user as we observe more data from them. With this strategy, data rich users will have large $\alpha$ and data poor users will have small $\alpha$. Thus data rich users will be penalized less for moving further away from the global state. This is close to treating our loss as the maximum a posteriori estimate of the users data distribution, as we observe more data. The rate of changing $\alpha$ could be chosen so as to minimize the loss of our approach on some held out user data, following the normal cross validation strategy for choosing hyperparameters. Alternatively, we may have domain specific intuition on how much personalization matters, and $\alpha$ provides an easy way to express this.

From the catastrophic forgetting perspective, our definition is similar to the work in \cite{aljundi2017memory}, which penalizes weights from moving away from what they were on a previous task. That work upweights the penalty for weights that have a high average gradient on the previous task, reasoning that such weights are likely to be most important. We directly penalize the loss of accuracy on other users, rather than indirectly penalizing that change, as the gradient based approach does. The indirect approach of \cite{aljundi2017memory} has the benefits of being scalable, as it may be expensive to recalculate total global loss and potentially adapting to unlabeled data. Still, we see a common motivation, as in both cases, we have some weighting for how much we want to allow our model to change in response to new examples. 

To calculate $\Loss \prn{D,\, M_i}$ we do not need to gather the data in a central location (which would violate our privacy constraint). It is enough to share $M_i$ with each other user, or some sampling of other users, and gather summary statistics of how well the model performs. We could then aggregate these summary statistics to evaluate how well $M_i$ does on $D$. However, sharing a user's model with other users still compromises the original user's privacy, since model weights potentially offer insight into user behavior. 

In practice, we often have a subset of user data that we can centralize from users who have opted in, or a large public curated dataset that is relevant to our task of interest. We treat such a dataset as a stand in for how users will generally behave. This approach does not compromise user privacy. Alternately, since our $\Loss \prn{D,\, M_i}$ is meant to regularize and stabilize our local models, there may be other approaches that achieve this global objective without directly measuring performance on global data. In future work, we will more deeply explore how best to measure this global loss without violating user privacy.

\section{Experiments}

We run an experiment with a simple model to demonstrate the trade-offs between personal and global performance and how the choice of $\alpha$ might affect the way we make future user predictions.

\subsection{Setup and Data}

We use the Stanford Sentiment Treebank (SSTB) \cite{socher2013recursive} dataset and evaluate how well we can learn models for sentiment. As a first step, we take the 200 most positive and 200 most negative words in the dataset, which we find by training a simple logistic regression model on the train set. We then run experiments simulating the existence of 2, 5, or 8 users. In each experiment, these words are randomly partitioned amongst users, and users are assigned sentences containing those words for their validation, train, and test sets. Sentences that contain no words in this top 400 are randomly assigned, and sentences that contain multiple of these are randomly assigned to one of the relevant users. This results in a split of the dataset in which each model has a subset of words that are significantly enriched for them, but are very underrepresented for all other models.

This split is meant to simulate a pathological case of user style; we try to simulate users in our train set that are very biased and almost non-overlapping in terms of the word choice they use to express sentiment. While this may not be the case for this specific review dataset, in general there will be natural language tasks in which users have specific slang, inside jokes, or acronyms that they use that may not be used by others. For such users, an ideal setup would adapt to their personal slang, while still leveraging global models to help understand the more common language they use.

\subsection{Architecture}

For each user we train completely separate models with the same architecture. Roughly following the baseline from the original SSTB paper, we classify sentences using a simple two-layer neural network, and use an average of word embeddings as input and a tanh non-linearity. We use 35 dimensional word embeddings, dropout of $0.5$ \cite{srivastava2014dropout}, and use ADAM to optimize \cite{kingma6980method}. We start with an initial learning rate of $0.001$, which we slowly decay, by multiplying by $0.95$, if the validation accuracy has not decreased after a fixed number of batches, which is the same across all experiments. Finally, we use early stopping in the case validation accuracy does not decrease after a fix number of batches, equivalent to 5 epochs on the full train set. Once trained we evaluate two ways of combining our fixed models, averaging model predictions, and simply taking the most confident models, where confidence is defined as the absolute difference between $0.5$ and the models prediction.  

\subsection{Evaluation Metrics}

To evaluate we use the train, validation, test splits as provided with the dataset, and use pytreebank \cite{pytreebank} to parse the data. We only evaluate on the sentence level data for test and validation sets. This model is not state of the art. However, we have experience putting models of similar size on low powered and memory constrained devices, and believe this model could realistically be deployed. Nevertheless, the model is sufficiently complicated to give us a sense of what happens as we try to combine separately trained models. In all tables, we report accuracy, and test accuracy for user specific data is evaluated solely on sentences that contain only their words and none of the other users' specific words. Global data scores represents the whole test set. We report all results averaged over 15 independent trials.

\section{Results and Analysis}

\subsection{Single User Performance on User-Specific Data Vs. Single User Performance on Global Data}

Unsurprisingly, as the second and third columns of Table \ref{tab:fE1_2a} show, single user models perform much better on their own heavily biased user-specific test set than on the global data. This makes sense as each model has purposely been trained on more words from their biased test set. Those words were also specifically selected to be polarizing, but the gap makes concrete the extent to which varying word usages can hurt model performance on this task. 

\begin{table}[ht]
\begin{tabular}{|p{1cm}|p{3.8cm}|p{3.4cm}|p{3.4cm}|p{4cm}|}
\hline
Num. Users & Single user model (user-specific dataset) & Single user model (global dataset) & Average aggregation (global dataset) & Confidence aggregation (global dataset) \\
\hline
2&0.826&0.797&0.816&0.803\\
\hline
3&0.824& 0.783&0.813&0.789\\
\hline
5&0.806&0.739&0.794&0.746 \\
\hline
8&0.795& 0.697& 0.772& 0.704\\
\hline
\end{tabular}
\caption{Accuracy by number of users. The second column reports accuracy of the single user model on the user-specific datasets. The last three columns report performance on the global dataset for the single user model and the two ensemble models.}\label{tab:fE1_2a}
\end{table}

\subsection{Single User Performance on User-Specific Data Vs. Ensembled Models on User-Specific Data}

As the number of users increases, the single user model outperforms both aggregation methods on user-specific data (Table \ref{tab:fE1_2b}). This is particularly pronounced for the confidence aggregation method: ensembling hurts performance across all experiments, with this effect increasing as we add users. As the number of users increases, for any given prediction we are less likely to choose the specific user's model, which performs best on their own dataset. The averaging aggregation method outperforms the confidence aggregation method and is competitive with the single user model for up to five users. However, for more than five users, the averaging approach starts to perform worse on the user's own data, again suggesting that we start to drown out much of the personal judgment and rely on global knowledge. 

\begin{table}[ht]
\begin{tabular}{|l|l|l|l|l|l|}
\hline
Num. Users& Difference (Average Aggregation) & Difference (Confidence Aggregation)\\
\hline
2&-0.001&0.013\\
\hline
3&-0.001&0.026\\
\hline
5&0.005&0.059\\
\hline
8&0.022&0.096\\

\hline
\end{tabular}
\caption{Comparison of ensemble model performance to single user model performance on user-specific data. ``Difference'' columns denote the single user model accuracy minus the ensemble model accuracy. As the number of users increases, user-specific models outperform ensemble models by increasingly wide margins.} \label{tab:fE1_2b}
\end{table}

\subsection{User Performance on Global Data Vs. Ensembled Models on Global Data}

While it might be easy to conclude that we should just use a single user model, Table \ref{tab:fE1_2d} demonstrates that the average-aggregated ensembled models outperform the single user model on global data, particularly as the number of users increases. Again, this is what we would expect, since the aggregated models have collectively been trained on more words in more examples, and ought to generalize better to unbiased and unseen data. This global knowledge is still important, as it may contain insights about phrases a user has only used a few times. This may be especially true for a user who has little data. Recall we divide the whole dataset amongst all users, so as the number of users increases, each user-specific model is trained on less data. In this case the lack of a word in the training set may not indicate that a user will never use that word. It may be that the user has not interacted with the system enough for their individual model to have fully learned their language.

\begin{table}[ht]
\begin{tabular}{|l|l|l|l|l|l|}
\hline
Num. Users& Difference (Average Aggregation) & Difference (Confidence Aggregation)\\
\hline
2&-0.019&-0.006\\
\hline
3&-0.029&-0.005\\
\hline
5&-0.055&-0.007\\
\hline
8&-0.075&-0.006\\
\hline
\end{tabular}
\caption{Comparison of ensemble model performance to single user model performance on global data. ``Difference'' columns denote single user model accuracy minus ensemble model accuracy. As the number of users increases, the average-aggregated ensemble model increasingly outperforms the single user model.} \label{tab:fE1_2d}
\end{table}

\subsection{Choosing an Approach Based on $\alpha$}

These experiments demonstrate the tensions between performing well on global and user data, the two terms in our loss in Equation \ref{eq:loss}. We can apply Equation \ref{eq:loss}, vary $\alpha$, and see at what point we should prefer different strategies.

Specifically, suppose we have two approaches we can choose from, with personalized losses of $p_0$ and $p_1$, and global losses of $g_0$ and $g_1$ respectively. If $\Loss_0 - \Loss_1 < 0$ the first approach is superior. We can solve for the $\alpha$ such that $\Loss_0 = \Loss_1$, where our loss term again comes from Equation \ref{eq:loss}. Plugging our definition in, we see that $$\alpha p_0 + (1-\alpha)  g_0 = \alpha p_1+ (1 - \alpha) g_1,$$ or equivalently, 
$$\alpha p_0 + (1 - \alpha) g_0 - \alpha p_1 - (1 - \alpha) g_1 = 0.$$ 
Rearranging this yields 
\begin{equation}\label{eq:alphaeq}
    \alpha = \frac{g_1 - g_0} {\prn{p_0 - p_1} - \prn{g_0 - g_1}},
\end{equation}
as our break even personalization point. For this value of $\alpha$, we ought to see our two models as equally valid solutions to the problem of personalization.

$\Loss_0 - \Loss_1$ is linear with respect to $\alpha$, so if $\Loss_0 - \Loss_1 \ge 0 $ for any $\alpha$ above our cutoff, it will be greater everywhere above the cutoff, and vice versa. This yields a rule for how to approach making a decision between multiple types of models. It only requires choosing a single hyperparameter $\alpha$ between 0 and 1, representing one's belief on how much personalization matters to the task at hand.

We illustrate how to apply these ideas to our experimental results. We see from Tables \ref{tab:fE1_2b} and \ref{tab:fE1_2d} that when we compare using a single model to averaging predictions from 5 models, we have: 

$$g_\text{average} - g_\text{single} = -0.054$$
$$p_\text{single} - p_\text{average} = -0.00523$$

So, our cutoff value of $\alpha$, where the single model and averaged models yield equivalent losses and we are indifferent between them, is $-0.0545 / (-0.0545 - 0.00523) = 0.9124$. We can also compute the ranges of $\alpha$ where we should prefer each model. Because, as explained above, $\Loss_\text{single} - \Loss_\text{average}$  is linear in $\alpha$, evaluating a single point above the cutoff suffices: we choose $\alpha = 1$ for computational convenience, and have $\Loss_\text{single} - \Loss_\text{average} = p_\text{single} - p_\text{average} = -0.00523 \le 0.0$. Consequently, we prefer the single model for values of $\alpha$ above the cutoff, and the averaged model for values of $\alpha$ below the cutoff. 

\section{Conclusion}
Our definition of personalization allows for a complete decoupling of models at train time, while only requiring aggregate knowledge of other models inference in order to potentially benefit from this global knowledge. In addition, it gives a practitioner a simple, one parameter way, of deciding how to choose amongst models that may have different strengths and weaknesses. Further, we have shown how this approach might look on a simplified dataset and model, and why the na\"ive approach of using a single model, or always aggregating all models, may sometimes not be optimal. 

In the future, we will work to develop better methods for combining this aggregate global knowledge, while not hurting user performance. To better protect user privacy, we will also consider alternate methods for regularizing our models outside of the global loss term, $\Loss \prn{D,\, M_i}$. We hope that this work will provide a useful framing for future work on personalization, learning in decentralized architectures, such as Ethereum and Bitcoin \cite{nakamoto2008bitcoin}, \cite{wood2014ethereum}, and serve as a guideline for situations in which the normal single loss and centralized server training paradigm cannot be used.

\bibliographystyle{amsplain}
\bibliography{ml}

\end{document}

%% file: MathMacros.tex
\newcommand{\sets}[1]{\left\{{#1}\right\}} 
\newcommand{\set}[2]{\sets{{#1}:\,{#2}}} 
\newcommand{\prn}[1]{\left({#1}\right)} 

\theoremstyle{plain}

\theoremstyle{definition}

\theoremstyle{remark}

\usepackage{enumerate}
\usepackage{amssymb}
\usepackage[mathscr]{eucal}
\usepackage{graphicx,color}


%% file: pai.bbl
\providecommand{\bysame}{\leavevmode\hbox to3em{\hrulefill}\thinspace}
\providecommand{\MR}{\relax\ifhmode\unskip\space\fi MR }
\providecommand{\MRhref}[2]{%
  \href{http://www.ams.org/mathscinet-getitem?mr=#1}{#2}
}
\providecommand{\href}[2]{#2}
\begin{thebibliography}{10}

\bibitem{abadi2016deep}
Mart{\'\i}n Abadi, Andy Chu, Ian Goodfellow, H~Brendan McMahan, Ilya Mironov,
  Kunal Talwar, and Li~Zhang, \emph{Deep learning with differential privacy},
  Proceedings of the 2016 ACM SIGSAC Conference on Computer and Communications
  Security, ACM, 2016, pp.~308--318.

\bibitem{al2016conversational}
Rami Al-Rfou, Marc Pickett, Javier Snaider, Yun-hsuan Sung, Brian Strope, and
  Ray Kurzweil, \emph{Conversational contextual cues: The case of
  personalization and history for response ranking}, arXiv preprint
  arXiv:1606.00372 (2016).

\bibitem{aljundi2017memory}
Rahaf Aljundi, Francesca Babiloni, Mohamed Elhoseiny, Marcus Rohrbach, and
  Tinne Tuytelaars, \emph{Memory aware synapses: Learning what (not) to
  forget}, arXiv preprint arXiv:1711.09601 (2017).

\bibitem{chen2017user}
Zheqian Chen, Ben Gao, Huimin Zhang, Zhou Zhao, Haifeng Liu, and Deng Cai,
  \emph{User personalized satisfaction prediction via multiple instance deep
  learning}, Proceedings of the 26th International Conference on World Wide
  Web, International World Wide Web Conferences Steering Committee, 2017,
  pp.~907--915.

\bibitem{cheng2016wide}
Heng-Tze Cheng, Levent Koc, Jeremiah Harmsen, Tal Shaked, Tushar Chandra,
  Hrishi Aradhye, Glen Anderson, Greg Corrado, Wei Chai, Mustafa Ispir, et~al.,
  \emph{Wide \& deep learning for recommender systems}, Proceedings of the 1st
  Workshop on Deep Learning for Recommender Systems, ACM, 2016, pp.~7--10.

\bibitem{deng2009imagenet}
Jia Deng, Wei Dong, Richard Socher, Li-Jia Li, Kai Li, and Li~Fei-Fei,
  \emph{Imagenet: A large-scale hierarchical image database}, Computer Vision
  and Pattern Recognition, 2009. CVPR 2009. IEEE Conference on, IEEE, 2009,
  pp.~248--255.

\bibitem{dwork2008differential}
Cynthia Dwork, \emph{Differential privacy: A survey of results}, International
  Conference on Theory and Applications of Models of Computation, Springer,
  2008, pp.~1--19.

\bibitem{fan2015bayesian}
Kai Fan, Allison~E Aiello, and Katherine~A Heller, \emph{Bayesian models for
  heterogeneous personalized health data}, arXiv preprint arXiv:1509.00110
  (2015).

\bibitem{french1999catastrophic}
Robert~M French, \emph{Catastrophic forgetting in connectionist networks},
  Trends in cognitive sciences \textbf{3} (1999), no.~4, 128--135.

\bibitem{kemker2017measuring}
Ronald Kemker, Angelina Abitino, Marc McClure, and Christopher Kanan,
  \emph{Measuring catastrophic forgetting in neural networks}, arXiv preprint
  arXiv:1708.02072 (2017).

\bibitem{kingma6980method}
Diederik~P Kingma and Jimmy~Ba Adam, \emph{A method for stochastic
  optimization. 2014}, arXiv preprint arXiv:1412.6980.

\bibitem{kirkpatrick2017overcoming}
James Kirkpatrick, Razvan Pascanu, Neil Rabinowitz, Joel Veness, Guillaume
  Desjardins, Andrei~A Rusu, Kieran Milan, John Quan, Tiago Ramalho, Agnieszka
  Grabska-Barwinska, et~al., \emph{Overcoming catastrophic forgetting in neural
  networks}, Proceedings of the National Academy of Sciences (2017), 201611835.

\bibitem{konevcny2016federated}
Jakub Kone{\v{c}}n{\`y}, H~Brendan McMahan, Daniel Ramage, and Peter
  Richt{\'a}rik, \emph{Federated optimization: distributed machine learning for
  on-device intelligence}, arXiv preprint arXiv:1610.02527 (2016).

\bibitem{konevcny2016federatedcommunication}
Jakub Kone{\v{c}}n{\`y}, H~Brendan McMahan, Felix~X Yu, Peter Richt{\'a}rik,
  Ananda~Theertha Suresh, and Dave Bacon, \emph{Federated learning: Strategies
  for improving communication efficiency}, arXiv preprint arXiv:1610.05492
  (2016).

\bibitem{lin2014microsoft}
Tsung-Yi Lin, Michael Maire, Serge Belongie, James Hays, Pietro Perona, Deva
  Ramanan, Piotr Doll{\'a}r, and C~Lawrence Zitnick, \emph{Microsoft coco:
  Common objects in context}, European conference on computer vision, Springer,
  2014, pp.~740--755.

\bibitem{malle2017more}
Bernd Malle, Nicola Giuliani, Peter Kieseberg, and Andreas Holzinger, \emph{The
  more the merrier-federated learning from local sphere recommendations},
  International Cross-Domain Conference for Machine Learning and Knowledge
  Extraction, Springer, 2017, pp.~367--373.

\bibitem{mcmahan2016federated}
H~Brendan McMahan, Eider Moore, Daniel Ramage, and Blaise~Aguera y~Arcas,
  \emph{Federated learning of deep networks using model averaging},  (2016).

\bibitem{nakamoto2008bitcoin}
Satoshi Nakamoto, \emph{Bitcoin: A peer-to-peer electronic cash system}, 2008.

\bibitem{CaliforniaPrivacy}
State of~California Department of Justice Office of~the Attorney~General,
  \emph{Privacy laws | state of california - department of justice - office of
  the attorney general}, \url{https://oag.ca.gov/privacy/privacy-laws},
  Accessed:2018-01-22.

\bibitem{park2017attend}
Cesc~Chunseong Park, Byeongchang Kim, and Gunhee Kim, \emph{Attend to you:
  Personalized image captioning with context sequence memory networks}, arXiv
  preprint arXiv:1704.06485 (2017).

\bibitem{GGDPR}
The~European Parliament and the Council Of The European~Union, \emph{Regulation
  (eu) 2016/679 of the european parliament and of the council of 27 april
  2016},
  \url{http://eur-lex.europa.eu/legal-content/EN/TXT/PDF/?uri=CELEX:32016R0679&from=en},
  2016, Accessed:2018-01-23.

\bibitem{pytreebank}
Jonathan Raiman, \emph{Stanford sentiment treebank loader in python},
  \url{https://github.com/JonathanRaiman/pytreebank}, Accessed:2018-01-05.

\bibitem{russakovsky2015imagenet}
Olga Russakovsky, Jia Deng, Hao Su, Jonathan Krause, Sanjeev Satheesh, Sean Ma,
  Zhiheng Huang, Andrej Karpathy, Aditya Khosla, Michael Bernstein, et~al.,
  \emph{Imagenet large scale visual recognition challenge}, International
  Journal of Computer Vision \textbf{115} (2015), no.~3, 211--252.

\bibitem{shi2017zhusuan}
Jiaxin Shi, Jianfei Chen, Jun Zhu, Shengyang Sun, Yucen Luo, Yihong Gu, and
  Yuhao Zhou, \emph{Zhusuan: A library for bayesian deep learning}, arXiv
  preprint arXiv:1709.05870 (2017).

\bibitem{smith2017federated}
Virginia Smith, Chao-Kai Chiang, Maziar Sanjabi, and Ameet Talwalkar,
  \emph{Federated multi-task learning}, arXiv preprint arXiv:1705.10467 (2017).

\bibitem{socher2013recursive}
Richard Socher, Alex Perelygin, Jean Wu, Jason Chuang, Christopher~D Manning,
  Andrew Ng, and Christopher Potts, \emph{Recursive deep models for semantic
  compositionality over a sentiment treebank}, Proceedings of the 2013
  conference on empirical methods in natural language processing, 2013,
  pp.~1631--1642.

\bibitem{srivastava2014dropout}
Nitish Srivastava, Geoffrey~E Hinton, Alex Krizhevsky, Ilya Sutskever, and
  Ruslan Salakhutdinov, \emph{Dropout: a simple way to prevent neural networks
  from overfitting.}, Journal of machine learning research \textbf{15} (2014),
  no.~1, 1929--1958.

\bibitem{wood2014ethereum}
Gavin Wood, \emph{Ethereum: A secure decentralised generalised transaction
  ledger}, Ethereum Project Yellow Paper \textbf{151} (2014).

\bibitem{zhang2007efficient}
Yi~Zhang and Jonathan Koren, \emph{Efficient bayesian hierarchical user
  modeling for recommendation system}, Proceedings of the 30th annual
  international ACM SIGIR conference on Research and development in information
  retrieval, ACM, 2007, pp.~47--54.

\bibitem{zigoris2006bayesian}
Philip Zigoris and Yi~Zhang, \emph{Bayesian adaptive user profiling with
  explicit \& implicit feedback}, Proceedings of the 15th ACM international
  conference on Information and knowledge management, ACM, 2006, pp.~397--404.

\end{thebibliography}
